\documentclass{article}

\usepackage[preprint,nonatbib]{neurips_2021}
\usepackage[numbers]{natbib}

\usepackage[utf8]{inputenc} 
\usepackage[T1]{fontenc}    
\usepackage{hyperref}       
\usepackage{url}            
\usepackage{booktabs}       
\usepackage{amsfonts}       
\usepackage{nicefrac}       
\usepackage{microtype}      
\usepackage{xcolor}         

\usepackage{arydshln}
\usepackage{placeins}
\usepackage{amsmath}
\usepackage{cleveref}
\usepackage{amssymb}
\usepackage{float}
\usepackage{printlen}
\usepackage{wrapfig}
\usepackage[bottom]{footmisc}
\usepackage{enumitem}
\usepackage{soul}
\usepackage{caption}
\usepackage{subcaption}
\usepackage{arydshln}
\usepackage{booktabs}
\usepackage[export]{adjustbox}
\usepackage{float}
\usepackage{mathtools}
\usepackage{multirow}
\usepackage{tabularx}

\newcommand{\R}{\mathbb{R}}
\newcommand{\model}{{MultiScaleGNN}}
\newcolumntype{Y}{>{\centering\arraybackslash}X}

\title{Simulating Continuum Mechanics with \\ Multi-Scale Graph Neural Networks}

\author{%
  Mario Lino \\
  Department of Aeronautics\\
  Imperial College London\\
  \texttt{mal1218@ic.ac.uk} \\
  \And
  Chris Cantwell \\
  Department of Aeronautics \\
  Imperial College London \\
  \And
  Anil A. Bharath \\
  Department of Bioengineering  \\
  Imperial College London \\
  \And
  Stathi Fotiadis \\
  Department of Bioengineering  \\
  Imperial College London \\
}

\begin{document}

\maketitle

\begin{abstract}
Continuum mechanics simulators, numerically solving one or more partial differential equations, are essential tools in many areas of science and engineering, but their performance often limits application in practice. Recent modern machine learning approaches have demonstrated their ability to accelerate spatio-temporal predictions, although, with only moderate accuracy in comparison.
Here we introduce \model, a novel multi-scale graph neural network model for learning to infer unsteady continuum mechanics.
\model \ represents the physical domain as an unstructured set of nodes, and it constructs one or more graphs, each of them encoding different scales of spatial resolution.
Successive learnt message passing between these graphs improves the ability of GNNs to capture and forecast the system state in problems encompassing a range of length scales.
Using graph representations, \model \ can impose periodic boundary conditions as an inductive bias on the edges in the graphs, and achieve independence to the nodes' positions.
We demonstrate this method on advection problems and incompressible fluid dynamics.
Our results show that the proposed model can generalise from uniform advection fields to high-gradient fields on complex domains at test time and infer long-term Navier-Stokes solutions within a range of Reynolds numbers.
Simulations obtained with \model \ are between two and four orders of magnitude faster than the ones on which it was trained.

\end{abstract}

\section{Introduction}
Forecasting the spatio-temporal mechanics of continuous systems is a common requirement in many areas of science and engineering.
Continuum mechanics models often describe the underlying physical laws by one or more partial differential equations (PDEs), whose complexity may preclude their analytic solution.
Numerical methods are well-established for approximately solving PDEs by discretising the the governing PDEs to form an algebraic system which can be solved \cite{spencer,nektar}.
Such solvers can achieve high levels of accuracy, but they are computationally expensive.
Deep learning techniques have recently been shown to accelerate physical simulations without significantly penalising accuracy.  
This has motivated an increasing interest in the use of deep neural networks (DNNs) to produce realistic real-time animations for computer graphics \cite{tompson2017accelerating,Kim2019,sanchez2020learning} and to speed-up simulations in engineering design and control \cite{Guo2016,Yilmaz2017,Zhang2018,Thuerey2018}.

Most of the recent work on deep learning to infer continuous physics  has focused on convolutional neural networks (CNNs) \cite{Wiewel2019,xiao2018novel,Lino2020}.
In part, the success of CNNs for these problems lies in their translation invariance and locality \cite{goodfellow2016deep}, which represent strong and desirable inductive biases for continuum-mechanics models.
However, CNNs constrain input and output fields to be defined on rectangular domains representated by regular grids, which is not suitable for more complex domains.
As for traditional numerical techniques, it is desirable to be able to adjust the resolution over space, devoting more effort where the physics are challenging to resolve, and less effort elsewhere.
An alternative approach to applying deep learning to geometrically and topologically complex domains are graph neural networks (GNNs), which can also be designed to satisfy spatial invariance and locality \cite{battaglia2018relational,wu2020comprehensive}.
In this paper we
describe a novel approach to applying GNNs for accurately forecasting the evolution of physical systems in complex and irregular domains.

\begin{figure}[ht]
\centering
\begin{tabular}{cc}
\includegraphics[clip, height=0.30\columnwidth, trim={0mm 0mm 0mm 0mm} ]{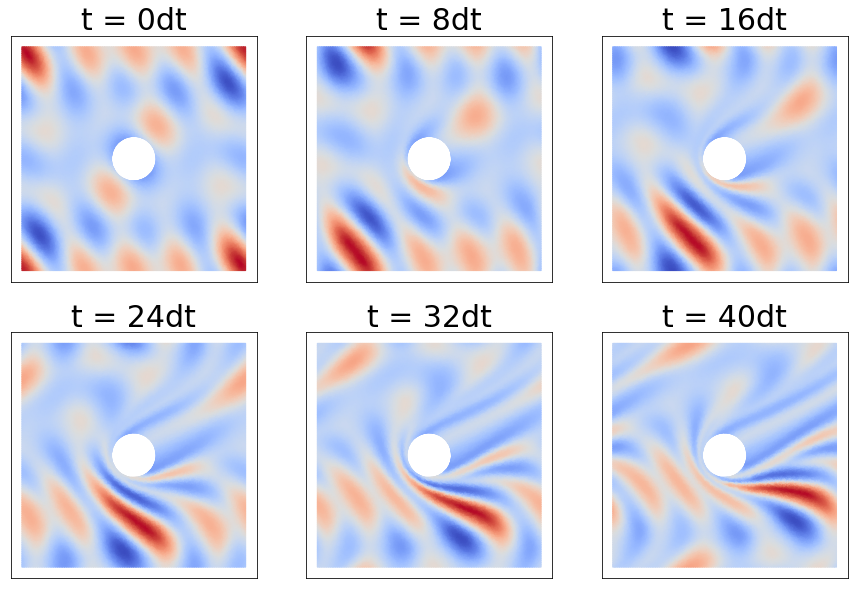}
&
\includegraphics[clip, height=0.30\columnwidth, trim={0mm 0mm 0mm 0mm}]{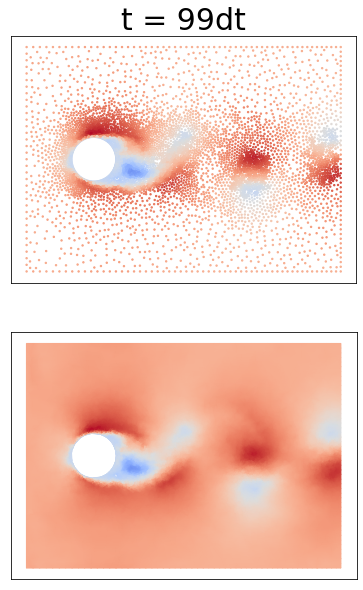} \\
{\small (a) Advection} &
{\small (b) Incompressible flow}
\end{tabular}
{\small \caption{\model \ forecasts continuous dynamics. We used it to simulate (a) advection
[\href{https://imgur.com/a/zdQHs5A}{\texttt{video}}]
and, (b) the incompressible flow around a circular cylinder
[\href{https://imgur.com/a/qY8mkpu}{\texttt{video}}]. 
In (a) the upper-lower and left-right boundaries are periodic.
In (b) the upper-lower boundaries are periodic. The upper figure shows the $u$ field predicted at a set of nodes obtained with adaptive coarsening/refinement, the bottom figure shows the interpolated field.
\label{fig:intro}
}}
\end{figure}

\paragraph{Contribution.}
We propose \model, a multi-scale GNN model, to forecast the spatio-temporal evolution of continuous systems discretised as unstructured sets of nodes. 
Each scale processes the information at different resolutions, enabling the network to more accurately and efficiently capture complex physical systems, such as fluid dynamics, with the same number of learnable parameters as a single level GNN.
We apply \model \ to simulate two unsteady fluid dynamics problems: advection and incompressible fluid flow around a circular cylinder within the time-periodic laminar vortex-shedding regime.
We evaluate the improvement in accuracy over long roll-outs as the number of scales in the model are increased.
Importantly, we show that the proposed model is independent of the spatial discretisation and the mean absolute error (MAE) decreases linearly as the distance between nodes is reduced.
We exploit this to apply adaptive graph coarsening and refinement.
We also demonstrate a technique to impose periodic boundary conditions as an inductive bias on the graph network connectivity.
\model \ simulations are between two and four orders of magnitude faster than the numerical solver used for generating the ground truth datasets, becoming a potential surrogate model for fast predictions.

\section{Background and related work}

\paragraph{Deep learning on graphs}
A directed graph, $G$, is defined as $G=(V,E)$, where $V=\{1, 2, \dots, |V|\}$ is a set of nodes and $E = \{1, 2, \dots, |E|\}$ is a set of directed edges, each of them connecting a sender node $s \in V$ to a receiver node $r \in V$.
Each node $i$ can be assigned a vector of node attributes $\mathbf{v}_i$, and each edge $k$ can be assigned a vector of edge attributes $\mathbf{e}_k$.
Gilmer, et al. (2017) \cite{Gilmer2017} introduced neural message passing (MP), a layer that updates the node attributes given the current node and edge attributes, as well as a set of learnable parameters.
Battaglia et al. (2018) \cite{battaglia2018relational} further generalised MP layers by introducing an edge-update function.
Their MP layer performs three steps: edge update, aggregation and node update \textemdash equations (\ref{eq:edge_model}), (\ref{eq:aggr}) and (\ref{eq:node_model}) respectively.

\begin{align}
        \mathbf{e}_k' &= f^e(\mathbf{e}_k,\mathbf{v}_{r_k},\mathbf{v}_{s_k}) \ \ \ \forall k \in E
        \label{eq:edge_model} \\
        \bar{\mathbf{e}}_r' &= \rho^{e \rightarrow v}(E_r') \ \ \ \ \ \ \ \ \ \ \ \ \  \forall r \in V
        \label{eq:aggr} \\
        \mathbf{v}_r' &= f^v(\bar{\mathbf{e}}_r', \mathbf{v}_r) \ \ \ \ \ \ \ \ \ \ \ \  \forall r \in V
        \label{eq:node_model}
\end{align}

Functions $f^e$ and $f^v$ are the edge and node update functions \textemdash typically multi-layer perceptrons (MLPs).
The aggregation function $\rho^{e \rightarrow v}$ is applied to each subset of updated edges going to node $r$, and it must be permutation invariant.
This MP layer has already proved successful in a range of data-driven physics problems \cite{sanchez2020learning,pfaff2020learning,battaglia2016interaction}.

\paragraph{CNNs for simulating continuous mechanics.}
During the last five years, most of the DNNs used for predicting continuous physics have included convolutional layers.
For instance, CNNs have been used to solve the Poisson's equation \citep{Tang2018,Ozbay2019}, to solve the steady N-S equations \citep{Guo2016,Yilmaz2017,Zhang2018,Thuerey2018,Miyanawala2017}, and to simulate unsteady fluid flows \citep{Kim2019,Wiewel2019,Lino2020,Lee2018,Sorteberg2018,Fotiadis2020}.
These CNN-based solvers are between one and four orders of magnitude faster than numerical solvers \citep{Guo2016,Lino2020}, and some of them have shown good extrapolation to unseen domain geometries and initial conditions \citep{Thuerey2018,Lino2020}.

\paragraph{Simulating physics with GNNs.}
Recently, GNNs have been used to simulate the motion of discrete systems of solid particles \cite{battaglia2016interaction,chang2016compositional,Sanchez-Gonzalez2018} and deformable solids and fluids discretised into lagrangian (or \textit{free}) particles \cite{sanchez2020learning,li2018learning,mrowca2018flexible,Li2019a}.
Battaglia, et al. (2016) \cite{battaglia2016interaction} and Chang, et al. (2016) \cite{chang2016compositional} represented systems of particles as graphs, whose nodes encode the state of the particles and whose edges encode their pairwise interactions; nodes and edges are individually updated through a single MP layer to determine the evolution of the system.
Of note is the work of Sanchez, et al. (2020), who discretised fluids into a set of free particles in order to apply the previous approach and infer fluid motion.
Further research in this area introduced more general MP layers \cite{Sanchez-Gonzalez2018,Li2019a,Mrowca2018}, high-order time-integration \cite{sanchez2019hamiltonian} and hierarchical models \citep{li2018learning,mrowca2018flexible}.
To the best of our knowledge Alet, et al. (2019) \cite{alet2019graph} were the first to explore the use of GNNs to infer Eulerian mechanics by solving the Poisson PDE. However, their domains remained simple, used coarse spatial discretisations and did not explore the generalisation of their model.
Later, Belbute-Peres, et al. (2020) \cite{belbute2020combining} introduced an hybrid model consisting of a numerical solver providing low-resolution solutions to the N-S equations and several MP layers performing super-resolution on them.
More closely related to our work, Pfaff, et al. (2020) \cite{pfaff2020learning} proposed a mesh-based GNN to simulate continuum mechanics, although they did not consider the use of MP at multiple scales of resolution.

\section{Model}
\subsection{Model definition}

\begin{figure}[ht]
\centering
\includegraphics[clip,width=0.9\columnwidth, trim={0mm 0mm 0mm 0mm}]{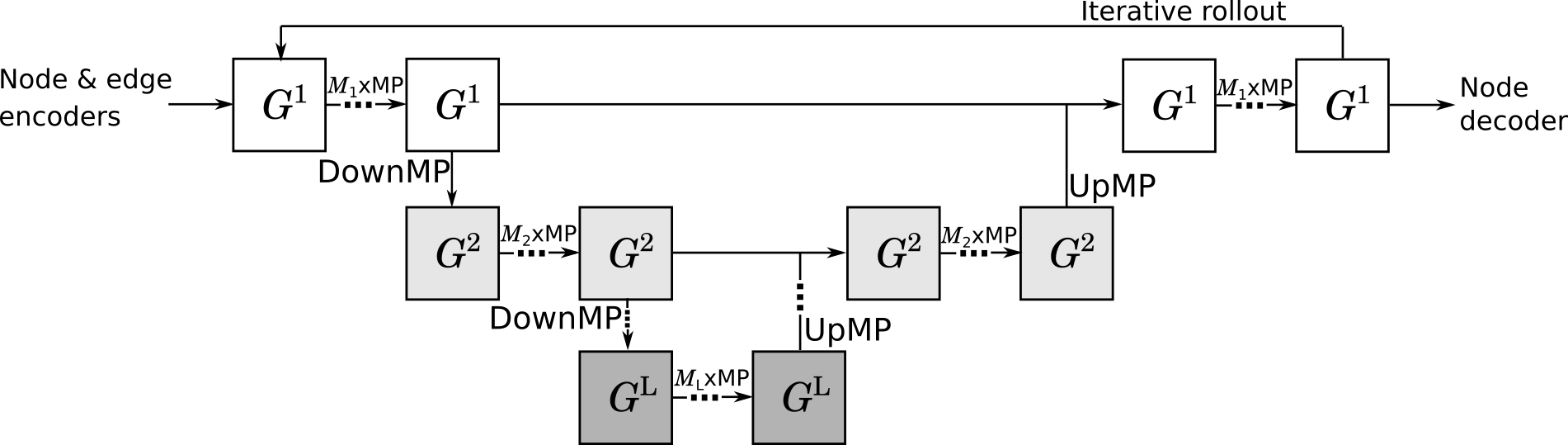}
{\small \caption{\model \ architecture. $G^1$ is an original high-resolution graph, $G^2$ is a lower-resolution graph obtained from $G^1$, $G^L$ is the lowest-resolution graph. $M_l$ MP layers are applied in each graph, and, DownMP and UpMP layers between them. 
\label{fig:net}
}}
\end{figure}

\model \ infers the temporal evolution of an $n$-dimensional field, $\mathbf{u}(t, \mathbf{x}) \in \R^{n}$, in a spatial domain $\mathcal{D} \subset \R^2$.\footnote{Model repository available on \textit{revealed on publication.}}
It requires the discretisation of $\mathcal{D}$ into a finite set of nodes $V^1$, with coordinates $\mathbf{x}^1_i \in \mathcal{D} $.
Given an input field $\mathbf{u}(t_0, \mathbf{x}_{V^1})$, at time $t=t_0$ and at the $V^1$ nodes, a single evaluation of
\model{} returns $\mathbf{u}(t_0+dt, \mathbf{x}_{V^1})$, where $dt$ is a fixed time-step size.
Each time-step is performed by applying MP layers in $L$ graphs and between them, as illustrated in Figure \ref{fig:net}.

The high-resolution graph $G^1$ consists of the set of nodes $V^1$
and a set of directed edges $E^1$ connecting these nodes.
In a complete graph there exist $|V^1|(|V^1|-1)$ edges, however, MP would be extremely expensive.
Instead, \model \ connects each node in $V^1$ (receiver node) to its $k$-closest nodes ($k$-sender nodes) using a k-nearest neighbors (k-NN) algorithm.
The node attributes of each node $\mathbf{v}^1_i$ are the concatenation of $\mathbf{u}(t_0, \mathbf{x}_{i})$, $\mathbf{p}_i$ and $\Omega_i$; where
$\mathbf{p}_i$ is a set of physical parameters at $\mathbf{x}_i$ (such as the Reynolds number in fluid dynamics) and
\begin{equation} \label{eq:omega}
\Omega_i := 
\begin{cases}
    1 & \text{if } \mathbf{x}_i \in  \partial_D \mathcal{D}, \\
    0 & \text{otherwise},
\end{cases}
\end{equation}
with $\partial_D\mathcal{D}$ representing the Dirichlet boundaries.
Each edge attribute $\mathbf{e}^1_k$ is assigned the relative position between sender node $s_k$ and receiver node $r_k$.
Node attributes and edge attributes are then encoded through two independent MLPs.

A MP layer applied to $G^1$ propagates the nodal and edge information only locally between connected (or \textit{neighbouring}) nodes.
Nevertheless, most continuous physical systems require this propagation at larger scales or even globally, for instance, in incompressible flows pressure changes propagate at infinite speed throughout the entire domain.
Propagating the nodal and edge attributes between nodes separated by hundreds of hops would necessitate an excessive number of MP layers, which is neither efficient nor scalable.
Instead, \model{} processes the information at $L$ scales, creating a graph for each level and propagating information between them in each pass.
The lower-resolution graphs ($G^2, G^3, \dots, G^L$; with $|V_1| > |V_2| > \dots > |V_L|$) possess fewer nodes and edges, and hence, a single MP layer can propagate attributes over longer distances more efficiently.
These graphs are described in section \ref{sec:multiscale_graphs}.
As depicted in Figure \ref{fig:net}, the information is first diffused and processed in the high-resolution graph $G^1$ through $M_1$ MP layers. It is then passed to $G^2$ through a downward MP (DownMP) layer.
In $G^2$ the attributes are again processed through $M_2$ MP layers.
This process is repeated $L-1$ times. The lowest resolution attributes (stored in $G^L$) are then passed back to the scale immediately above through an upward message passing (UpMP) layer.
Attributes are successively passed through $M_l$ MP layers at scale $l$ and an UpMP layer from scale $l$ to scale $l-1$ until the information is ultimately processed in $G^1$.
Finally, a MLP decodes the nodal information to return the predicted field at time $t_0 + dt$ at the $V^1$ nodes.
To apply MP in the $L$ graphs, \model \ uses the MP layer summarised in equations (\ref{eq:edge_model}) to (\ref{eq:node_model}), with the mean as the aggregation function.

\subsection{Multi-scale graphs}
\label{sec:multiscale_graphs}
Each graph $G^l=(V^l, E^l)$ with $l = 2, 3, \dots, L$; is obtained from graph $G^{l-1}$ by first dividing $\mathcal{D}$ into a regular grid with cell size $d_x^l \times d_y^l$ (Figure \ref{fig:pool}a exemplifies this for $l=2$).
For each cell, provided that there is at least one node from $V^{l-1}$ on it, a node is added to $V^l$.
The nodes from $V^{l-1}$ on a given cell $i$ and the node from $V^{l}$ on the same cell are denotes as child nodes, $Ch(i) \subset V^{l-1}$, and parent node, $i \in V^l$, respectively.
The coordinates of each parent node is the mean position of its children nodes.
Each edge $k \in E^{l}$ connects sender node $s_k \in V^{l}$ to receiver node $r_k \in V^l$, provided that there exists at least one edge from $Ch(s_k)$ to $Ch(r_k)$ in $E^{l-1}$.
Edge $k$ is assigned the mean edge attribute of the set of edges going from $Ch(s_k)$ to $Ch(r_k)$.

\begin{small}
\begin{figure}[ht]
\centering
\begin{tabular}{cc}
\includegraphics[clip, height=0.23\columnwidth, trim={0mm, -50mm, 0mm, 0mm}]{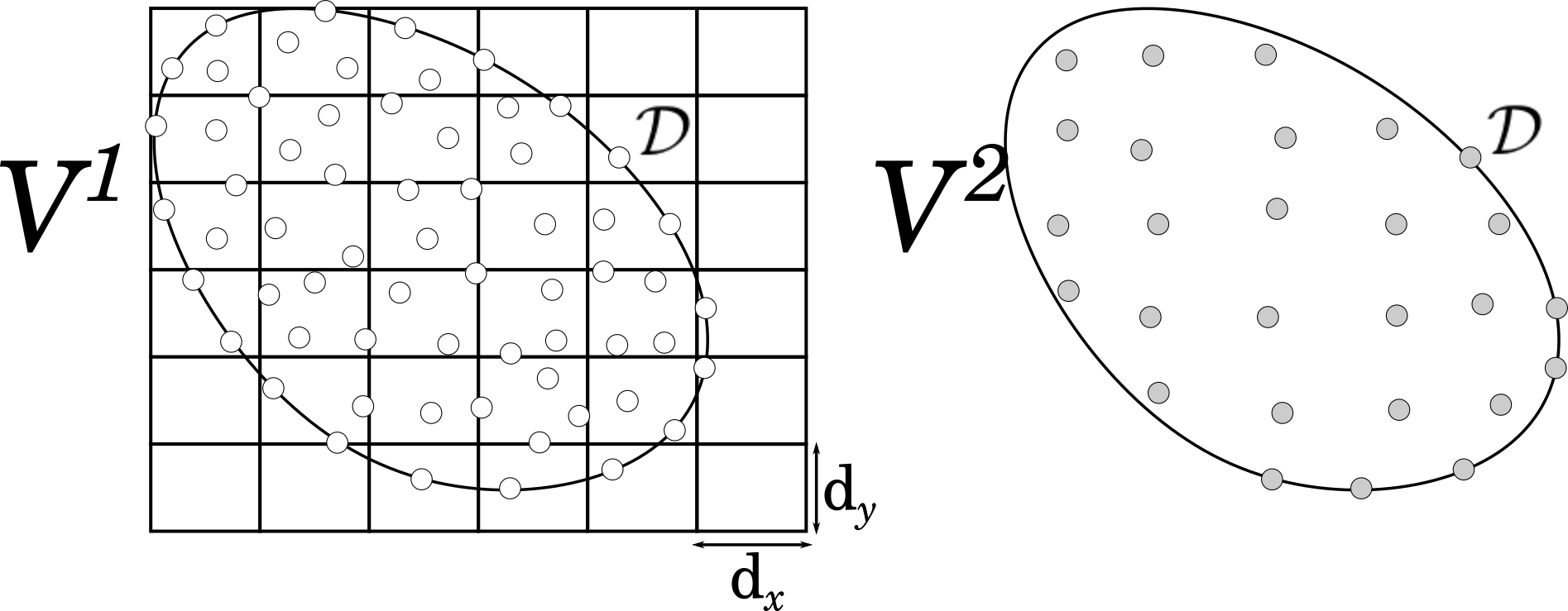} &
\includegraphics[height=0.25\columnwidth]{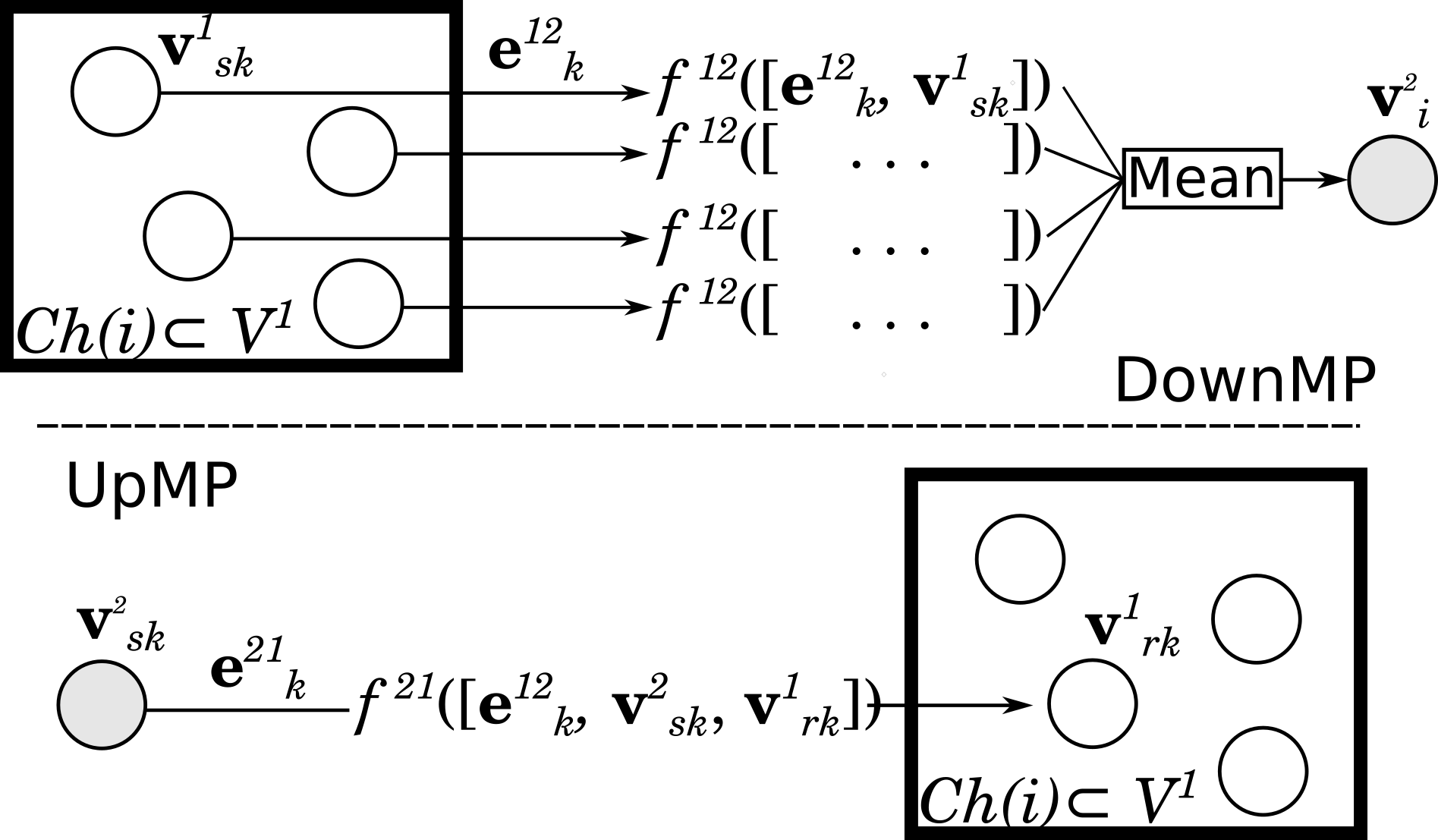} \\
{\small (a) Building $V^2$ from $V^1$} &
{\small (b) DownMP and UpMP layers}
\end{tabular}
{\small \caption{(a) $V^2$ is obtained from $V^1$ by partitioning $\mathcal{D}$ using a grid with cell size $d_x^2 \times d_y^2$ and assigning a parent node $i \in V^2$ at the mean position of the child nodes $Ch(i) \subset V^{1}$ in each cell. (b) DownMP and UpMP diagram.
\label{fig:pool}
}}
\end{figure}
\end{small}

\paragraph{Downward message-passing (DownMP).}
To perform MP from $G^{l-1}$ to $G^l$ (see Figure \ref{fig:pool}b), a set of directed edges, $E^{l-1,l}$, is created.
Each edge $k \in E^{l-1,l}$ connects node $s_k \in V^{l-1}$ to its parent node $r_k \in V^{l}$, with edge attributes assigned as the relative position between child and parent nodes.
A DownMP layer applies a common edge-update function, $f^{l-1,l}$, to edge $k$ and node $s_k$.
It then assigns to each node attribute $\mathbf{v}^l_i$ the mean updated attribute of all the edges in $E^{l-1,l}$ arriving to node $i$, i.e.,

\begin{equation}
    \mathbf{v}^l_i = \frac{1}{|Ch(i)|} \sum_{k: r_k=i} f^{l-1, l}([\mathbf{e}^{l-1, l}_k,  \mathbf{v}^{l-1}_{s_k}]), \ \ \ \forall i \in 1, \ldots, |V^{l}|.
    \label{eq:pool}
\end{equation}

\paragraph{Upward message-passing (UpMP).}
To pass and process the node attributes from $G^{l+1}$ to $G^{l}$, \model \ defines a set of directed edges, $E^{l+1, l}$.
These edges are the same as in $E^{l,l+1}$, but with opposite direction.
An UpMP layer applies a common edge-update function, $f^{l+1,l}$, to each edge $k \in E^{l+1,l}$ and both its sender (at scale $l+1$) and receiver (at scale $l$) nodes, directly updating the node attributes in $G^{l}$, i.e.,
\begin{equation}
    \mathbf{v}^{l}_{r_k} = f^{l+1,l}([\mathbf{e}^{l+1,l}_k, \mathbf{v}^{l+1}_{s_k}, \mathbf{v}^{l}_{r_k}]), \ \ \ \forall k \in E^{l+1,l}.
    \label{eq:unpool}
\end{equation}
UpMP layers leave the edge attributes of $G^{l}$ unaltered.
To model functions $f^{l-1,l}$ and $f^{l+1,l}$ we use MLPs.

\subsection{Periodic boundaries} \label{sec:bc}

\begin{wrapfigure}{R}{0.5\textwidth}
\begin{center}
\includegraphics[width=0.5\columnwidth]{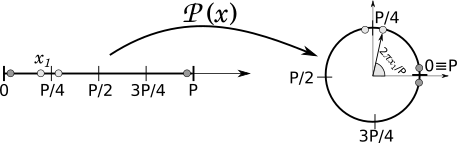}
\end{center}
{\small \caption{Function $\mathcal{P}$ transforms a periodic direction into a circumference with unitary radius.
\label{fig:periodic}
}}
\end{wrapfigure}

MultiLevelGNN imposes periodicity along $x$ and $y$ directions as an inductive bias on $E^1$.
For this purpose, each periodic direction is transformed through a function $\mathcal{P} : \R \mapsto \R^2$ into a circumference with unitary radius,
\begin{equation}
    \mathcal{P}(x) := [\cos(2\pi x/P), \sin(2\pi x/P)],
\end{equation}
where $P$ is the period along the spatial coordinate under consideration ($x$ or $y$ direction). The k-NN algorithm that creates $E^1$ (see Figure \ref{fig:periodic}) is then applied to the transformed coordinates.
Even though $E^1$ is created based on nodal proximity in the transformed space, the edge attributes are still assigned the relative positions between nodes in the original coordinate system, except at the periodic boundaries where $P$ is added or subtracted to ensure a consistent representation.

\subsection{Model training} \label{sec:model_train}

\model \ was trained by supervising at each time-point the loss $\mathcal{L}$ between the predicted field $\hat{\mathbf{u}}(t, \mathbf{x}^1_{V^1})$ and the ground-truth field $\mathbf{u}(t, \mathbf{x}^1_{V^1})$ obtained with a high-order finite element solver.
$\mathcal{L}$ includes the mean-squared error (MSE) at the $V^1$ nodes, the MAE at the $V^1$ nodes on Dirichlet-boundaries, and the MSE for the rate of change along the $E^1$ edges.
$\mathcal{L}$ can be expressed as
\begin{align}
    \mathcal{L} = \mathrm{MSE}&\Big(\hat{\mathbf{u}}(t, \mathbf{x}^1_{V^1}),\mathbf{u}(t, \mathbf{x}^1_{V^1}) \Big) \nonumber\\
    &+ \lambda_d\, \mathrm{MAE}\Big(\hat{\mathbf{u}}(t, {\mathbf{x}^1_{V^1} \in \partial_D\mathcal{D}}),\mathbf{u}(t, \mathbf{x}^1_{V^1} \in \partial_D\mathcal{D}) \Big) \nonumber\\
    &+ \frac{\lambda_e}{|E^1|} \sum_{\forall k \in E^1} \mathrm{MSE}\bigg(\frac{\hat{\mathbf{u}}(t, \mathbf{x}^1_{r_k}) - \hat{\mathbf{u}}(t, \mathbf{x}^1_{s_k})}{||\mathbf{e}_k||_2}, \frac{\mathbf{u}(t, \mathbf{x}^1_{r_k}) - \mathbf{u}(t, \mathbf{x}^1_{s_k})}{||\mathbf{e}_k||_2}\bigg).
\label{eq:loss}
\end{align}
The hyper-parameters $\lambda_d$ and $\lambda_e$ are selected based on the physics to be learnt. Training details for our experiments can be found in Appendix \ref{sec:model_detail}.

\section{Training datasets}
We generated datasets to train \model \ to simulate advection and incompressible fluid dynamics.\footnote{Datasets available on \textit{revealed on publication}.}

\paragraph{Advection.}
Datasets \texttt{AdvBox} and \texttt{AdvInBox} contain simulations of a scalar field advected under a uniform velocity field (with magnitude in the range 0 to 1) on a square domain ($[0,1]\times[0,1]$) and a rectangular domain ($[0,1]\times[0,0.5]$) respectively.
\texttt{AdvBox} domains have periodic conditions on all four boundaries, whereas \texttt{AdvInBox} domains have upper and lower periodic boundaries, a prescribed Dirichlet condition on the left boundary, and a zero-Neumann condition on the right boundary.
The initial states at $t_0$ are derived from two-dimensional truncated Fourier series with random coefficients and a random number of terms.
For each dataset, a new set of nodes $V^1$ is generated at the beginning of every training epoch.
For advection models, $\mathbf{u}(t,\mathbf{x}_i) \in \R$ is the advected field and $\mathbf{p}_i(\mathbf{x}_i) \in \R^2$ are the two components the velocity field.

\paragraph{Incompressible fluid dynamics.}
Dataset \texttt{NS} contains simulations of the periodic vortex shedding around a circular cylinder at Reynolds number $\mathrm{Re} \in [500,1000]$.
The upper and lower boundaries are periodic, hence, the physical problem is equivalent to an infinite vertical array of cylinders.
The distance between cylinders, $H$, is randomly sampled between $4$ and $6$.  
Each domain is discretised into approximately 7000 nodes.
For flow models, $\mathbf{u}(t,\mathbf{x}_i) \in \R^3$ contains the velocity and pressure fields and $\mathbf{p}_i \in \R$ is the Reynolds number.
Further details of the training and testing datasets are included in Appendix \ref{sec:data_detail}.

\section{Results and discussion}
We analyse the generalisation to unseen physical parameters, the effect of the number of scales and the independence to the spatial discretisation.
All the results, except Tables \ref{table:mae_adv_scales} and \ref{table:mae_ns_scales}, were obtained with 3-scale models.
Model details and hyper-parameters are included in Appendix \ref{sec:model_detail}.\footnote{Videos comparing the ground truth and the predictions can be found on \\ \href{https://imperialcollegelondon.box.com/s/f6eqb25rt14mhacaysqn436g7bup3onz}{\texttt{https://imperialcollegelondon.box.com/s/f6eqb25rt14mhacaysqn436g7bup3onz}}}.

\begin{figure}[ht]
\centering
\begin{tabular}{cc}
\includegraphics[clip, height=0.37\columnwidth, trim={0mm, 0mm, 0mm, 0mm}]{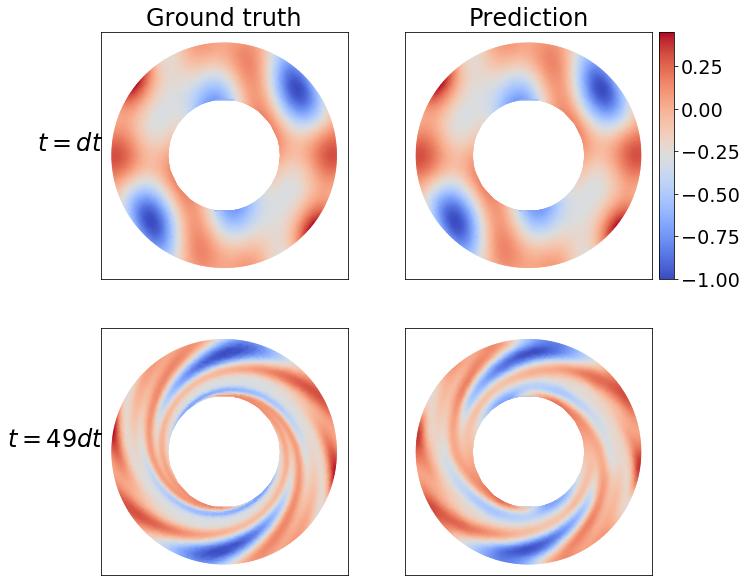} &
\includegraphics[clip, height=0.37\columnwidth, trim={0mm, 0mm, 0mm, 0mm}]{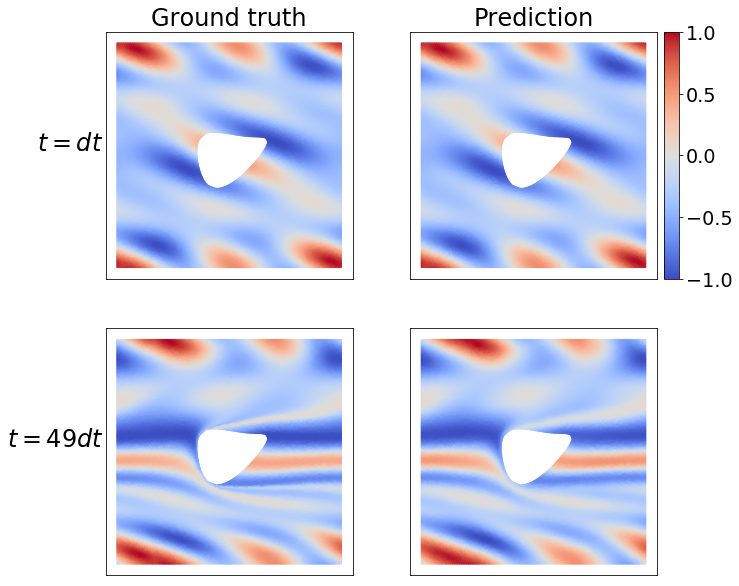} \\
{\small (a) Advection in a Taylor-Couette flow} &
{\small (b) Advection around a body made of splines}
\end{tabular}
{\small \caption{Ground truth and \model \ predictions for 
(a) the advection of scalar field in a Taylor-Couette flow [\href{https://imgur.com/a/q99ArnK}{\texttt{video}}], and 
(b) the advection around a body made of splines with periodicity along $x$ and $y$ [\href{https://imgur.com/a/NSTIS4T}{\texttt{video}}].
\label{fig:taylor_splines}
}}
\end{figure}

\paragraph{Generalisation.}
We first consider the \model{} model trained to infer advection.
Despite \model{} being trained on square and rectangular domains and uniform velocity fields, it generalises to complex domains and non-uniform velocity fields (obtained using the N-S equations with $Re=1$).
As an example of a closed domain, we consider the Taylor-Couette flow in which the inner and outer boundaries rotate at different speeds.
Figure \ref{fig:taylor_splines}a shows the ground truth and predictions for a simulation where the inner wall is moving faster than the outer wall in an
anti-clockwise direction.
In the Taylor-Couette flow, the structures present in the advected field increase in spatial frequency with time due to the shear flow.
This challenges both the networks ability to capture accurate solutions as well as its ability to generalise.
After 49 time-steps, the model maintains high accuracy in transporting both the lower and the higher frequencies.
We attribute this to the model's ability to process both high and low resolutions and the frequency range included in the training datasets \textemdash training datasets missing high-frequency structures resulted in a considerably higher diffusion.

We also evaluate the predictions of \model{} on open domains containing obstacles of different shapes (circles, squares, ellipses and closed spline curves).
Figure \ref{fig:taylor_splines}b shows the ground truth and predictions for a field advected around a body made of arbitrary splines, with upper-lower and left-right periodic boundaries.
The predictions show a strong resemblance to the ground truth, although after 49 time-steps we note that the predicted fields begin to show a higher degree of diffusion in the wake behind the obstacle.

\begin{small}
\begin{figure}[ht]
\centering
\begin{tabular}{clcl}
    \raisebox{1ex-\height}{(a)} &
    \raisebox{1ex-\height}{
        \includegraphics[clip, width=0.41\textwidth, trim=30mm 0mm 125mm 0mm ]{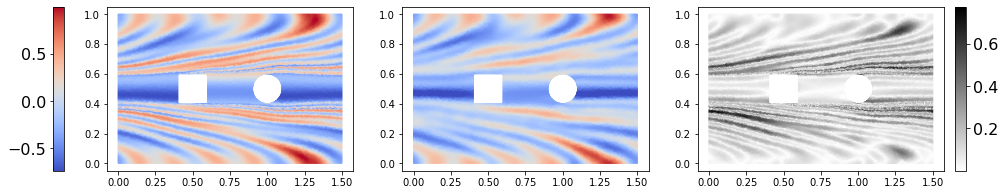}
    } &
    \raisebox{1ex-\height}{(b)} &
    \raisebox{1ex-\height}{
        \includegraphics[clip, width=0.41\textwidth, trim=30mm 0mm 125mm 0mm ]{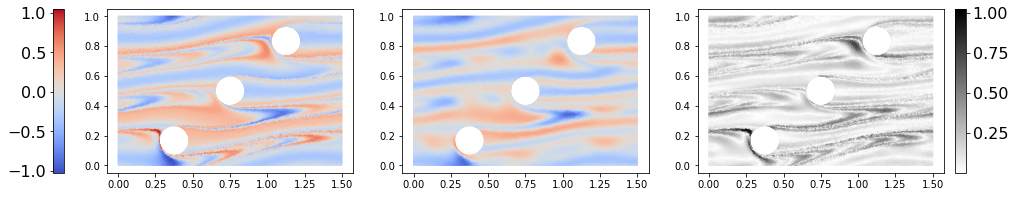}
    }
\end{tabular}
\caption{Ground truth and \model \ predictions for the advected field after 99 time-steps pass
(a) squared and circular obstacles [\href{https://imgur.com/a/CNawqxm}{\texttt{video}}], 
(b) circular obstacles [\href{https://imgur.com/a/jQBoR7z}{\texttt{video}}]. Upper-lower and left-right boundaries are periodic.}
\label{fig:long}
\end{figure}
\end{small}

We also assess the quality of long-term inference.
Figures \ref{fig:long}a and \ref{fig:long}b show the ground truth and predictions after 99 time-steps for two obstacles with periodicity in the $x$ and $y$ directions.
It can be seen that \model{} successfully generalises to infer advection around multiple obstacles.
However, the MAE increased approximately linearly every time-step.
This limitation could be explained by the simplicity of the training datasets, and it could be mitigated by training on non-uniform velocity fields.

\begin{figure}[ht]
\centering
\includegraphics[clip,width=0.9\columnwidth, trim={0mm 65mm 0mm 0mm}]{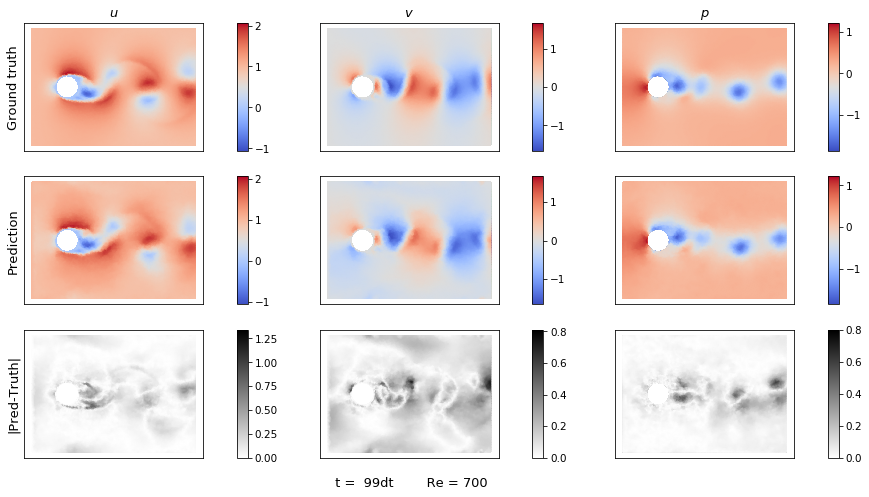}
{\small \caption{Ground truth and \model \ predictions for the horizontal velocity field, $u$, vertical velocity field, $v$, and pressure field, $p$, after 99 time-steps for $\mathrm{Re}=700$ [\href{https://imgur.com/a/vWX2DrB}{\texttt{video}}]. 
\label{fig:uvp}
}}
\end{figure}

\begin{figure}[ht]
\centering
\begin{tabular}{cc}
\includegraphics[clip, height=0.3\columnwidth, trim={0mm, 0mm, 0mm, 0mm}]{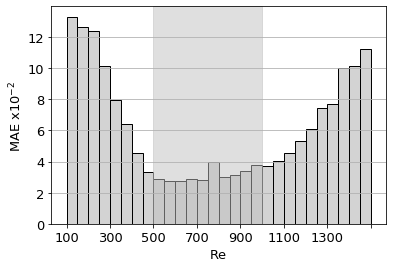} &
\includegraphics[clip, height=0.3\columnwidth, trim={0mm, 0mm, 0mm, 0mm}]{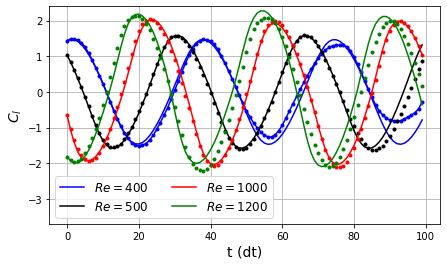} \\
{\small (a) Reynolds vs MAE } &
{\small (b) Lift coefficient}
\end{tabular}
{\small \caption{
(a) MAE over 99 time-points for different Reynolds numbers. The grey region indicates the interpolation region.
(b) Lift coefficient for $\mathrm{Re}=400, 500, 1000, 1200$. The continuos lines are the ground truth, whereas the dotted lines are the predictions.
\label{fig:ns_gen}
}}
\end{figure}

\model{} was also trained to simulate unsteady incompressible fluid dynamics in a range of Reynolds numbers between 500 and 1000.
It shows very good interpolation to unseen Reynolds numbers within this range.
For instance, Figure \ref{fig:uvp} compares the ground truth and predicted fields after 99 time-steps for $\mathrm{Re}=700$.
The MAE increases for Reynolds numbers lower than 400 and higher than 1200, as illustrated in Figure \ref{fig:ns_gen}a, which represents the MAE for a range of Reynolds numbers between 100 and 1500.
Figure \ref{fig:ns_gen}b shows the temporal evolution of the lift coefficient (vertical component of the integral of the pressure forces on the cylinder walls) for simulations with $\mathrm{Re}=400, 500, 1000$ and $1200$ (with $H=5$).
It can be noticed that for $\mathrm{Re}=500$ and $1000$, both at the edge of our training range, the ground truth (continuous lines) and the predictions (dotted lines) match very well over the entire simulation time.
For $\mathrm{Re}=400$ and $1200$ the lift predictions are qualitatively correct but begin to differ slightly in amplitude and phase.
The reason for the limited extrapolation may be the complexity of the N-S equations, which result into shorter wakes and higher frequencies as the Reynolds number is increased.

\paragraph{Multiple scales.}
We evaluate the accuracy of \model{} with $L=1,2,3$ and $4$; the architectural details of each model are included in Appendix \ref{sec:model_detail}.
Tables \ref{table:mae_adv_scales} and \ref{table:mae_ns_scales} collect the MAE for the last time-point and the mean of all the time-points on the testing datasets.
Incompressible fluids have a global behaviour since pressure waves travel at an infinite speed.
The addition of coarser layers helps the network learn this characteristic and achieve significantly lower errors.
For dataset \texttt{NSMidRe}, within the training regime of $\mathrm{Re}$, and \texttt{NSHighRe} there is a clear benefit from using four scales.
Interestingly, dataset \texttt{NSLowRe} shows a clear gain only when using two levels. This may be in part due to this dataset being in the extrapolation regime of $\mathrm{Re}$.

In contrast, for the advection datasets, the MAE does not substantially decrease when using an increasing number of levels.
In advection, information is propagated only locally and at a finite speed.
However, an accurate advection model must guarantee that it propagates the nodal information at least as fast as the velocity field driving the advection, similar to the Courant–Friedrichs–Lewy (CFL) condition used in numerical analysis.
In \model, the coarser graphs help to satisfy this condition, while the original graph maintains a detailed representation of structures in the advected fields.
Thus, GNNs for simulating both global and local continuum physics can benefit from learning information at multiple scales.
As a comparison to Pfaff, et al. (2020) \cite{pfaff2020learning}, a GNN with 16 sequential MP-layers (GN-Blocks) results in a MAE of $5.852 \times 10^{-2}$ on our \texttt{NSMidRe} dataset; whereas \model{} with the same number and type of MP layers, but distributed among 3 scales, results in a lower MAE of $3.201\times 10^{-2}$.

\begin{small}
\begin{table}[ht]
\centering
\caption{MAE $\times 10^{-2}$ on the advection testing sets for \model \ models with $L = 1, 2, 3, 4$} \label{table:mae_adv_scales}
\begin{tabularx}{\textwidth}{lYY|YY|YY|YY} 
\toprule
\multirow{2}{*}{Datasets} & \multicolumn{2}{c}{$L=1$} & \multicolumn{2}{c}{$L=2$} & \multicolumn{2}{c}{$L=3$} & \multicolumn{2}{c}{$L=4$} \\

                  &
Step 49 & All &
Step 49 & All &
Step 49 & All &
Step 49 & All \\

\midrule

\texttt{AdvTaylor} & 7.794 & 3.661 & \textbf{7.055} & \textbf{3.157} & 7.424 & 3.355 & 7.626 & 3.441 \\

\texttt{AdvCircle} & 3.508 & 1.758 & 3.119 & 1.547 & \textbf{2.986} & \textbf{1.475} & 3.486 & 1.629 \\

\texttt{AdvCircleAng} & 3.880 & 1.923 & 3.618 & 1.762 & \textbf{3.464} & \textbf{1.689} & 3.493 & 1.690 \\

\texttt{AdvSquare} & 3.765 & 1.853 & 3.493 & 1.689 & 3.279 & 1.571 & \textbf{3.219} & \textbf{1.556} \\

\texttt{AdvEllipseH} & 3.508 & 1.758 & 3.119 & 1.547 & \textbf{2.986} & \textbf{1.475} & 3.486 & 1.629 \\

\texttt{AdvEllipseV} & 3.770 & 1.867 & 3.748 & 1.853 & 3.450 & 1.708 & \textbf{3.334} & \textbf{1.653} \\

\texttt{AdvSplines} & 4.021 & 1.957 & 3.960 & 1.876 & 3.720 & 1.758 & \textbf{3.599} & \textbf{1.705} \\

\texttt{AdvInCir} & 9.886 & 6.008 & 10.755 & 6.298 & 14.948 & 8.491 & \textbf{9.670} & \textbf{5.808} \\

\bottomrule
\end{tabularx}
\label{tab:adv_scales}
\end{table}
\end{small}

\begin{small}
\begin{table}[ht]
\centering
\caption{MAE $\times 10^{-2}$ on the N-S testing sets for \model \ models with $L = 1, 2, 3, 4$} \label{table:mae_ns_scales}
\begin{tabularx}{\textwidth}{lYY|YY|YY|YY} 
\toprule
\multirow{2}{*}{Datasets} & \multicolumn{2}{c}{$L=1$} & \multicolumn{2}{c}{$L=2$} & \multicolumn{2}{c}{$L=3$} & \multicolumn{2}{c}{$L=4$} \\

                  &
Step 99 & All &
Step 99 & All &
Step 99 & All &
Step 99 & All \\

\midrule

\texttt{NSMidRe} & 13.028 & 8.033 & 4.731 & 3.666 & 3.95 & 3.201 & \textbf{3.693} & \textbf{3.032}\\

\texttt{NSLowRe} & 12.056 & 8.193 & \textbf{9.756} & \textbf{6.861} & 11.058 & 7.466 & 11.556 & 7.603 \\

\texttt{NSHighRe} & 16.176 & 11.42 & 9.949 & 7.385 & \textbf{8.936} & 7.17 & 9.085 & \textbf{6.822} \\

\bottomrule
\end{tabularx}
\end{table}
\end{small}

\begin{figure}[ht]
\begin{small}
\centering
\begin{tabular}{llll}
    \raisebox{-\height}{(a)} &
    \raisebox{-\height}{
        \includegraphics[clip, height=0.28\textwidth, trim=0mm 0mm 0mm 0mm ]{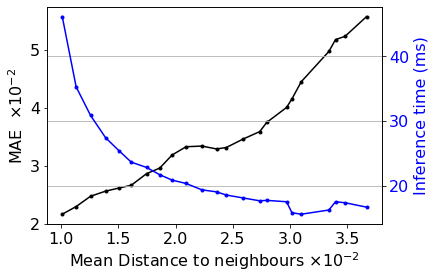}
    } &
    \raisebox{-\height}{(b)} &
    \raisebox{-\height}{
        \includegraphics[clip, height=0.28\textwidth]{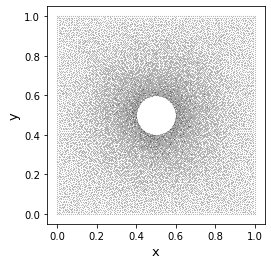}
    }
\end{tabular}
\caption{(a) For sample in Figure \ref{fig:intro}a., the MAE over 49 time-points and the inference time per time-step for a number of quasi-uniform discretisations.
(b) $V^1$ nodes used for the simulation shown in Figure \ref{fig:intro}a.}
\label{fig:edge}
\end{small}
\end{figure}

\paragraph{Discretisation independence.}
\model{} processes the relative position between neighbouring nodes, but it does not consider the absolute position of nodes.
It was also trained with random node distributions for each training example.
This ensures \model{} is independent to the set of nodes $V^1$ chosen to discretise the physical domain.
To demonstrate this, we consider a simulation from the \texttt{AdvCircleAng} dataset (see Figure \ref{fig:intro}a).
Figure \ref{fig:edge}a shows the MAE for a number of $V^1$ sets, with nodes evenly distributed on the domain.
It can be seen that the MAE decreases linearly as the mean \textit{distance-to-neighbour} is reduced -- at least within the interval shown.
Figure \ref{fig:edge}a also shows how the inference time per time-step increases exponentially as the spatial resolution is increased.
For this reason, the $V^1$ sets we feed to \model \ include a higher node count around solid walls.
As an example, the nodes $V^1$ used for the simulation depicted in Figure \ref{fig:intro}a are represented in Figure \ref{fig:edge}b.

Adaptive remeshing is a technique used by numerical solvers to ensure sufficient resolution is present where needed in the computational domain.
It dynamically modifies the spatial discretisation at every time-step.
Likewise, we implemented an algorithm to locally increase or decrease the spatial resolution of $V^1$ before every model evaluation.
Our algorithm increases the node count where the gradients of $\mathbf{u}$ are higher and decreases it where these are lower. 
Figure \ref{fig:intro}b shows the $u$ field predicted by \model \ combined with this technique.
It can be seen that the resolution is reduced far from the wake and increased close to the cylinder walls and on the vortices travelling downstream.
Adaptive coarsening/refinement allowed us to achieved an 11\% reduction of the MAE at the 99$^{\text{th}}$ time-point on the \texttt{NSMidRe} dataset.
More details are included in Appendix \ref{sec:adaptive}.

\paragraph{Computational efficiency.} \label{sec:performance}
We compare \model \ and the spectral/$hp$ element numerical solver \cite{nektar} used for generating the training and testing datasets.
For advection, \model \ simulations are one order of magnitude faster when they run on a CPU, and almost three orders of magnitude faster on a GPU.
The inference time per time-step for \texttt{AdvTaylor} is around 30 ms, and around 90 ms for the remaining testing datasets on a Tesla T4 GPU.
For the N-S equations, \model \ achieves an speed-up of three orders of magnitude running on a CPU, and four orders of magnitude on a Tesla T4 GPU; with an inference time per time-step of 50 ms on a Tesla T4 GPU.

\section{Conclusion}
\model \ is a novel multi-scale GNN model for inferring mechanics on continuous systems discretised into unstructured sets of nodes.
Unstructured discretisations allow complex domains to be accurately represented and the node count to be adjusted over space. Multiple coarser levels allow high and low-resolution mechanics to be efficiently captured.
For spatially local problems, such as advection, MP layers in coarse graphs guarantee that the information is diffused at the correct speed while MP layers in finer graphs provide sharp resolution of the fields.
In global and local problems, such as incompressible fluid dynamics, 
the coarser graphs are particularly advantageous, since they enable global characteristics such as pressure waves to be effectively learnt.
\model \ interpolates to unseen spatial discretisations of the physical domains, allowing it to adopt efficient discretisations and to dynamically and locally modify them to further improve the accuracy.
\model \ also generalises to advection on complex domains and velocity fields and it interpolates to unseen Reynolds numbers in fluid dynamics.
Inference is between two and four orders of magnitude faster than with the high-order solver used for generating the training datasets.
This work is a significant advancement in the design of flexible, accurate and efficient neural simulators.

\FloatBarrier
\bibliographystyle{unsrt}
\bibliography{bib}

\appendix

\section{Datasets details} \label{sec:data_detail}

\subsection{Advection datasets} \label{sec:adv_datasets}

\begin{figure}[ht]
\begin{small}
\centering
\begin{tabular}{llll}
    \raisebox{-\height}{(a)} &
    \raisebox{-\height}{
        \includegraphics[clip, height=0.25\textwidth, trim=0mm 0mm 0mm 0mm ]{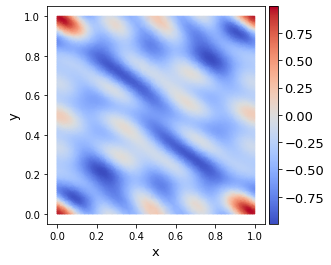}
    } &
    \raisebox{-\height}{(b)} &
    \raisebox{-\height}{
        \includegraphics[clip, height=0.25\textwidth, trim=0mm 0mm 0mm 0mm ]{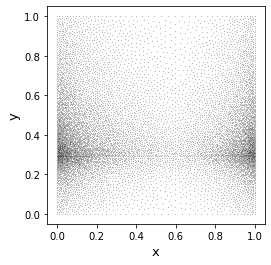}
    }
\end{tabular}
\caption{(a) Initial condition with with $M=5$ and $N=7$ from the \texttt{AdvBox} dataset. (b) One of the sets of nodes $V^1$ used during training. \label{fig:pbox_example}}
\end{small}
\end{figure}

We solved the two-dimensional advection equation using Nektar++, an spectral/hp element solver  \cite{spencer,nektar}. The advection equation reads
\begin{equation}
    \frac{\partial \varphi}{\partial t} + u(x,y) \frac{\partial \varphi}{\partial x} + v(x,y) \frac{\partial \varphi}{\partial y} = 0, 
\label{eq:adv}
\end{equation}
where $\varphi(t,x,y)$ is the advected field, and, $u$ and $v$ are the horizontal and vertical components of the velocity field respectively.
As initial condition, $\varphi_0$, we take an scalar field derived from a two-dimensional Fourier series with $M \times N$ random coefficients, specifically
\begin{gather}
    \varphi_0 = \sum_{m=0}^M \sum_{n=0}^N c_{m,n} \varphi_{m,n} \cdot \exp \Big(-2 (x-x_c)^2 -2 (x-x_c)^2 \Big), \label{eq:ic} \\
    \text{with} \ \ \ \varphi_{m,n} = \Re \bigg \{ \exp\Big(i 2 \pi (mx+ny) \Big)  \bigg \}. \label{eq:phi_mn}
\end{gather}
Coefficients $c_{m,n}$ are sampled from a uniform distribution between 0 and 1, and, integers $M$ and $N$ are randomly selected between 3 and 8.
In equation (\ref{eq:phi_mn}), $x_c$ and $y_c$ are the coordinates of the centre of the domain. 
The initial field $\varphi_0$ is scaled to have a maximum equal to $1$ and a minimum equal to $-1$.
Figure \ref{fig:pbox_example}a shows an example of $\varphi_0$ with $M=5$ and $N=7$
on a squared domain.
We created training and testing datasets containing advection simulations with 
50 time-points each, equispaced by a time-step size $dt = 0.03$. 
A summary of these datasets can be found in Table \ref{table:adv_datasets}.

\paragraph{Training datasets.}
We generated two training datasets: \texttt{AdvBox} with 1500 simulations and \texttt{AdvInBox} with 3000 simulations.
In these datasets we impose a uniform velocity fields with random values for $u$ and $v$, but constrained to $u^2+v^2 \leq 1$.
In dataset \texttt{AdvBox} the domain is a square ($\mathbf{x} \in [0,1]\times[0,1]$) with periodicity in $x$ and $y$.
In dataset \texttt{AdvInBox} the domain is a rectangle ($\mathbf{x} \in [0,1]\times[0,0.5]$) with periodicity in $y$, a Dirichlet condition on the left boundary and a homogeneous Neumann condition on the right boundary -- as an additional constraint, $u \geq 0$.
During training, a new set of nodes $V^1$ is selected at the beginning of every epoch.
The node count was varied smoothly across the different regions of the domains, as illustrated in Figure \ref{fig:pbox_example}b.
The sets of node were created with Gmsh, a finite-element mesher. 
The \textit{element size} parameter was set to 0.012 in the corners and the centre of the training domains, and set to $\sqrt{10}$ and $1/\sqrt{10}$ times that value at one random control point on each boundary. 
The mean number of nodes in the \texttt{AdvBox} and \texttt{AdvInBox} datasets are 98002 and 5009 respectively. 

\begin{figure}[h]
\begin{small}
    \centering
    \begin{adjustbox}{minipage=\linewidth,scale=0.9}
    \centering
    \begin{tabular}{cccccc}
        \raisebox{-\height}{(a)} &
        \raisebox{-\height}{
        \includegraphics[width=0.2\textwidth]{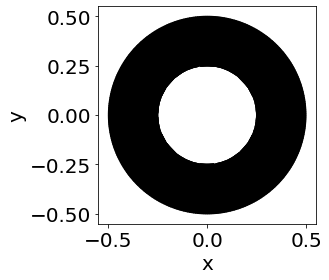}}
        &
        \raisebox{-\height}{(b)} &
        \raisebox{-\height}{
        \includegraphics[width=0.2\textwidth]{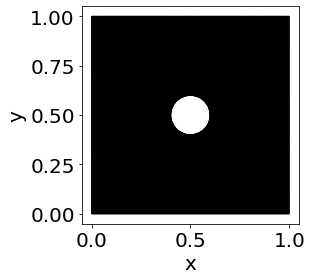}}
        &
        \raisebox{-\height}{(c)} &
        \raisebox{-\height}{
        \includegraphics[width=0.2\textwidth]{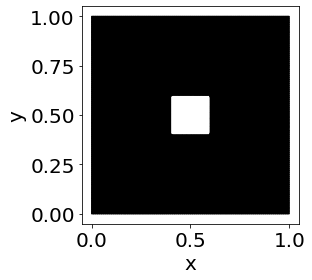}}
        \\
        \raisebox{-\height}{(d)} &
        \raisebox{-\height}{
        \includegraphics[width=0.2\textwidth]{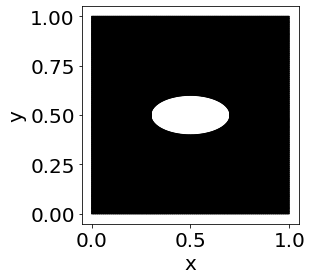}}
        &
        \raisebox{-\height}{(e)} &
        \raisebox{-\height}{
        \includegraphics[width=0.2\textwidth]{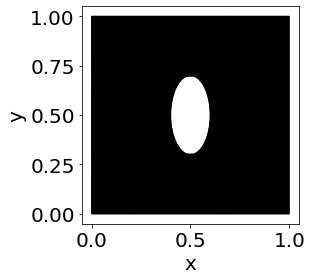}}
        &
        \raisebox{-\height}{(f)} &
        \raisebox{-\height}{
        \includegraphics[width=0.2\textwidth]{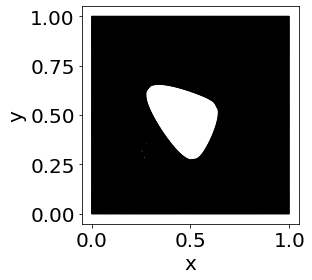}}
    \end{tabular}
    \end{adjustbox}
    \caption{Physical domains (black areas) on our testing datasets.}
    \label{fig:geo}
\end{small}
\end{figure}

\paragraph{Testing datasets.} 
We generated eight testing datasets, each of them containing 200 simulations.
These datasets consider advection on more complex open and closed domains with non-uniform velocity fields.
The domains employed are represented in Figure \ref{fig:geo}, and the testing datasets are listed in Table \ref{table:adv_datasets}.
The velocity fields were obtained from the steady incompressible Navier-Stokes equations with $Re=1$.
In dataset \texttt{AdvTaylor} the inner and outer walls spin at a velocity randomly sampled between $-1$ and $1$.
In datasets \texttt{AdvCircle}, \texttt{AdvSquare}, \texttt{AdvEllipseH}, \texttt{AdvEllipseV} and \texttt{AdvSplines} there is periodicity along $x$ and $y$, and a horizontal flow rate between $0.2$ and $0.75$ is imposed.
The obstacles inside the domains on the \texttt{AdvSplines} dataset are made of closed spline curves defined from six random points.
Dataset \texttt{AdvCircleAng} is similar to \texttt{AdvCircle}, but the flow rate forms an angle between $-45 \deg$ and $45 \deg$ with the $x$ axis.
The domain in dataset \texttt{AdvInCir} has periodicity along $y$, a Dirichlet condition on the left boundary (with $0.2 \leq u^2 + v^2  \leq 0.75$ and $-45 \deg  \leq \arctan(v/u)  \leq 45 \deg$), and a homogeneous Neumann condition on the right boundary.
The set of nodes $V^1$ were generated using Gmsh with an element size equal to 0.005 on the walls of the obstacles and 0.01 on the remaining boudaries.

\begin{table}[ht]
\centering
\caption{Advection training and testing datasets}
\label{table:adv_datasets}
\begin{tabularx}{\textwidth}{lcccc} 
\toprule
Dataset                 & Flow type                      & Domain  & \#Nodes               & Train/Test   \\ 
\midrule
\texttt{AdvBox}      & Open, periodic in $x$ and $y$ & $[0,1] \times [0,1]$     & 9601-10003            & Training \\
\texttt{AdvInBox}      & Open, periodic in $y$ & $[0,1] \times [0,0.5]$     & 4894-5135            & Training \\
\texttt{AdvTaylor}      & Closed, Taylor-Couette flow             & Figure \ref{fig:geo}a    & 7207           & Testing \\
\texttt{AdvCircle}      & Open, periodic in $x$ and $y$ & Figure \ref{fig:geo}b     & 19862            & Testing \\
\texttt{AdvCircleAng}   & Open, periodic in $x$ and $y$ & Figure \ref{fig:geo}b     & 19862           & Testing \\
\texttt{AdvSquare}      & Open, periodic in $x$ and $y$ & Figure \ref{fig:geo}c     & 19956            & Testing \\
\texttt{AdvEllipseH}    & Open, periodic in $x$ and $y$ & Figure \ref{fig:geo}d     & 20210            & Testing \\
\texttt{AdvEllipseV}    & Open, periodic in $x$ and $y$ & Figure \ref{fig:geo}e     & 20221           & Testing \\
\texttt{AdvSplines}     & Open, periodic in $x$ and $y$ & Figure \ref{fig:geo}f     & 19316-20389            & Testing \\
\texttt{AdvIncir}       & Open, periodic in $y$         & Figure \ref{fig:geo}b     & 19862           & Testing \\
\bottomrule
\end{tabularx}
\end{table}

\subsection{Incompressible fluid dynamics datasets} \label{sec:ns_datasets}
We solved the two-dimensional incompressible Navier-Stokes equation using the high-order solver Nektar++.
The Navier-Stokes equations read
\begin{gather}
        \frac{\partial u}{\partial x} + \frac{\partial v}{\partial y} = 0, \\
        \frac{\partial u}{\partial t} + u \frac{\partial u}{\partial x} + v\frac{\partial u}{\partial y} = - \frac{\partial p}{\partial x} + \frac{1}{Re}\bigg( \frac{\partial^2 u}{\partial x^2} + \frac{\partial^2 u}{\partial y^2} \bigg), \\
        \frac{\partial v}{\partial t} + u \frac{\partial v}{\partial x} + v\frac{\partial v}{\partial y} = - \frac{\partial p}{\partial y} + \frac{1}{Re}\bigg( \frac{\partial^2 v}{\partial x^2} + \frac{\partial^2 v}{\partial y^2} \bigg), 
\label{eq:ns}
\end{gather}
where $u(t,x,y)$ and $v(t,x,y)$ are the horizontal and vertical components of the velocity field, $p(t,x,y)$ is the pressure field, and $Re$ is the Reynolds number.
We consider the flow around an infinite vertical array of circular cylinder, with diameter $D=1$, equispaced a distance $H$ randomly sampled between $4D$ and $6D$.
The width of the domain is $7D$ and the cylinders axis is at $1.5D$ from the left boundary.
The left boundary is an inlet with $u = 1$, $v=0$ and $\partial p/ \partial x = 0$; the right boundary is an outlet with $\partial u/ \partial x = 0$, $\partial v/ \partial x=0$ and $p = 0$; and, the cylinder walls have a no-slip condition. 
The numerical solutions obtained for this flow at different Reynolds numbers are well known \cite{jiang2021large}.
In our simulations we select $Re$ values that yield solutions in the laminar vortex-shedding regime, and we only include the periodic stage.
The sets of nodes $V^1$ employed for each simulation were created using Gmsh placing more nodes around the cylinders walls (see Figure \ref{fig:nodes}a).
The mean number of nodes in these sets is 7143.
Each simulation contains $100$ time-points equispaced by a time-step size $dt=0.1$.
The training and testing datasets are listed in Table \ref{table:ns_datasets}.

\begin{table}[ht]
\centering
\caption{Incompressible flow training and testing datasets}
\label{table:ns_datasets}
\begin{tabular}{lccc} 
\toprule
Dataset                 & $Re$ & \#Simulations    & Train/Test   \\ 
\midrule
\texttt{NS}      & 500-1000 & 1000  & Training \\
\texttt{NSMidRe} & 500-1000 & 250 & Testing \\
\texttt{NSLowRe} & 100-500 & 250 & Testing \\
\texttt{NSMidRe} & 1000-1500 & 250 & Testing \\
\bottomrule
\end{tabular}
\end{table}

\section{Model details} \label{sec:model_detail}

\paragraph{Hyper-parameters choice.}
\model's hyper-parameters have different values for advection models and fluid dynamics models, Table \ref{table:hyper} collects the values we employed.
All MLPs use SELU activation functions \cite{klambauer2017self}, and, batch normalisation \cite{ba2016layer}.
Our results (except results in Tables \ref{table:mae_adv_scales} and \ref{table:mae_ns_scales}) were obtained with $L=3$, and, $M_1=2$, $M_2=2$, $M_3=4$ for advection; and $M_1=4$, $M_2=3$, $M_3=2$ for fluid dynamics.

\begin{table}[ht]
\centering
\caption{\model \ details}
\label{table:hyper}
\begin{tabular}{lcc} 
\toprule
Hyper-parameters                 & Advection & Fluid dynamics   \\ 
\midrule
$k$ & 4 & 8 \\
$d_x^2, d_y^2$ & 0.02 & 0.15 \\
$d_x^3, d_y^3$ & 0.04 & 0.30 \\
$d_x^4, d_y^4$ & 0.08 & 0.60 \\
\#Layers in edge encoder & 2 & 2 \\
\#Layers in node encoder & 2 & 2 \\
\#Layers in node decoder & 2 & 2 \\
\#Layers in edge MLPs & 2 & 3 \\
\#Layers in node MLPs & 2 & 2 \\
\#Layers in DownMP MLPs & 2 & 3 \\
\#Layers in UpMP MLPs & 2 & 3 \\
\#Neurons per layer & 128 & 128 \\
\bottomrule
\end{tabular}
\end{table}

\paragraph{Multi-scale experiments.}
The number of MP layers at each scale used for obtaining the results in Tables \ref{table:mae_adv_scales} and \ref{table:mae_ns_scales} are listed in the table below.
Notice that the total number of MP layers at each scale $l<L$, is $2\times M_l$.

\begin{table}[ht]
\centering
\caption{$M_l$ in the multi-scale experiments}
\begin{tabular}{lcc} 
\toprule
$L$                 & Advection & Fluid dynamics   \\ 
\midrule
$L=1$   & $M_1=4$  & $M_1=8$  \\
$L=2$   & $M_1=2, M_2=4$  & $M_1=4, M_2 = 4$  \\
$L=3$   & $M_1=2, M_2=2, M_3 = 4 $  & $M_1=4, M_2 = 2, M_3 = 4$  \\
$L=4$   & $M_1=2, M_2=2, M_3 = 2, M_4=4 $  & $M_1=4, M_2 = 2, M_3 = 2, M_4=4$  \\
\bottomrule
\end{tabular}
\end{table}

\paragraph{Training details.}
We trained \model \ models on a internal cluster using 4 CPUs, 86GB of memory, and a RTX6000 GPU with 24GB.
We fed 8 graphs per batch.
First, each training iteration predicted a single time-point, and, every time the training loss decreased below a threshold (0.01 for advection and 0.005 for fluid dynamics) we increased the number of iterative time-steps by one, up to a limit of 10.
We used the loss function given by equation (\ref{eq:loss}) with $\lambda_d=0.25$, and, $\lambda_e=0.5$ for advection and $\lambda_e=0$ for fluid dynamics.
The initial time-point was randomly selected for each prediction, and, we added to the initial field noise following a uniform distribution between -0.01 and 0.01. 
After each time-step, the models' weights were updated using the Adam optimiser with its standard parameters \cite{kingma2014adam}.
The learning rate was set to $10^{-4}$ and multiplied by 0.5 when the training loss did not decrease after five consecutive epochs, also, we applied gradient clipping to keep the Frobenius norm of the weights gradients below or equal to one.

\section{Adaptive coarsening/refinement} \label{sec:adaptive}

\begin{figure}[ht]
\centering
\begin{tabular}{ccc}
\includegraphics[clip, height=0.26\columnwidth, trim={0mm 0mm 0mm 0mm} ]{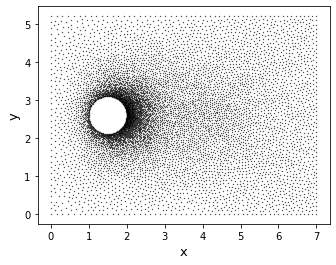}
&
\includegraphics[clip, height=0.26\columnwidth, trim={0mm 0mm 0mm 0mm}]{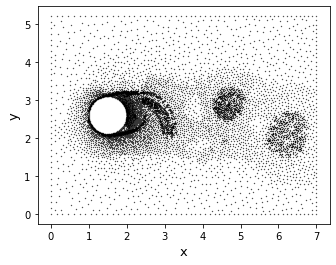}
&
\includegraphics[clip, height=0.26\columnwidth, trim={0mm 0mm 0mm 0mm}]{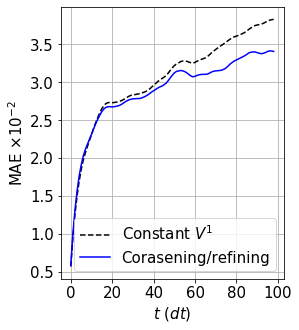} \\
{\small (a) Original $V^1$} &
{\small (b) Coarsened/refined $V^1$ at time $t$} &
{\small (c) MAE}
\end{tabular}
{\small \caption{(a) Original set of nodes and (b) set of nodes after applying our coarsening/refinement algorithm with $c=0.4$ and $f=0.1$.
(c) MAE on the \texttt{NSMidRe} dataset for \model \ without adaptive coarsening/refinement and \model \ combined with our coarsening/refinement algorithm.
\label{fig:nodes}
}}
\end{figure}

We implemented an algorithm to locally coarsen and refine the sets of nodes inputted to \model \ based on the gradients of the fields at these nodes.
The input to the coarsening/refinement algorithm is a set of nodes $V^1$, and the output is the same set with some of its nodes removed and some new nodes added to it.
The coarsening happens in regions where the amount of nodes is more than enough to capture the spatial variations of the input field, whereas the refinement happens in regions where more resolution is required.
The algorithm begins by connecting each node to its $k$-closest nodes, we denote the resulting set of edges by $E^{c/f}$.
Then, $\overline{\Delta u}$ is computed at each node.
We define $\overline{\Delta u}$ as
\begin{align}
        \Delta u_k & := || \mathbf{u}(t, \mathbf{x}_{r_k}) - \mathbf{u}(t, \mathbf{x}_{s_k}) ||_2 \ \ \ \forall k \in E^{c/f},
        \label{eq:c/f_1} \\
        \overline{\Delta u}_i & := \sum_{k:r_k=i}  \Delta u_k \ \ \ \ \  \ \  \ \  \ \ \ \ \ \ \ \ \ \ \ \ \ \ \ \  \forall i \in V^1,
        \label{eq:c/f_2}
\end{align}
and, it summarises the mean magnitude of the gradients of field $\mathbf{u}$ at each node.

The set of nodes $V^1_c$, which contains the nodes with a value of $\overline{\Delta u}$ below the $c |V^1|^{\text{th}}$-lowest value, is applied a coarsening algorithm; whereas the set of nodes $V^1_f$, which contains the nodes with a value above the $f |V^1|^{\text{th}}$-highest, is applied a refinement algorithm.
Values $c$ and $f$ can be chosen depending on the performance requirements.
The coarsening is based on Guillard's coarsening \cite{guillard1993node}, and, it consists on removing the $k$-closest nodes to node $i \in V^1_c$ provided that node $i$ has not been removed yet.
The refinement algorithm we implemented can be interpreted as an inverse of Guillard's coarsening: for each node $i \in V^1_f$, $k$ triangles with node $i$ and its $k$-closest nodes at their vertices are created, then, new nodes are added at the centroids of those triangles.
Figure \ref{fig:nodes}a shows an example of a set of nodes provided to our coarsening/refinement algorithm, and Figure \ref{fig:nodes}b shows the resulting set of nodes for $c=0.4$ and $f=0.1$.
Notice that, in line with Guillard's coarsening, the nodes on the boundaries of the domain are not considered by our algorithm.
We trained \model \ on the \texttt{NS} dataset with sets of nodes obtained for $c$ between 0.3 and 0.5, and $f$ between 0.05 and 0.15.
Then, during inference time, the sets of nodes $V^1$ fed to \model \ are coarsened/refined before every time-step.
Figure \ref{fig:nodes}c compares the MAE on the \texttt{NSMidRe} dataset for \model \ without adaptive coarsening/refinement and \model \ combined with our coarsening/refinement algorithm.
For a fair comparison, adaptive-\model \ predictions are interpolated to the primitive set of nodes, and the MAEs for both models are computed in the same set of nodes.
From this figure, it is clear, that having the ability to dynamically adapt the resolution can help to further improve the long-term accuracy.

\end{document}